# Question Answering in Natural Language: the Special Case of Temporal Expressions

**Armand Stricker**
LISN-CNRS, Université Paris-Saclay
armand.stricker@universite-paris-saclay.fr

**Abstract**

Although general question answering has been well explored in recent years, temporal question answering is a task which has not received as much focus. Our work aims to leverage a popular approach used for general question answering, answer extraction, in order to find answers to temporal questions within a paragraph. To train our model, we propose a new dataset, inspired by SQuAD, specifically tailored to provide rich temporal information. We chose to adapt the corpus WikiWars, which contains several documents on history's greatest conflicts. Our evaluation shows that a deep learning model trained to perform pattern matching, often used in general question answering, can be adapted to temporal question answering, if we accept to ask questions whose answers must be directly present within a text.

## 1 Introduction

Question answering is an automatic language processing task that aims to search for information in a text or database to answer a question in natural language. It is a task that differs from the query of a search engine, because it aims to exempt the user from querying the data using a formal language query. This type of method is particularly useful when the database is very large or poorly documented, or when the textual data to be queried is difficult to structure.

General question answering is a subject that has already been widely explored, and we have therefore decided to focus on a particular part of this research domain. The purpose of this paper will be to explore question answering as it relates to temporal information in English texts.

This is a task that can vary in difficulty: the temporal structure and the amount of temporal data can fluctuate quite a lot depending on the text, which can therefore be relatively simple to analyze or relatively complex, depending also on the questions asked. A set of clear definitions is therefore imperative.

### 1.1 Definitions

First, we need to define and limit what constitutes a *temporal expression*. Temporal information is most often expressed through a phrase or expression that describes a point in time or duration.

For this work, we define a temporal expression (timex) as any expression that denotes a moment or interval, or any other temporal reference that is not based on an event. Indeed, although an event can be located in time, it does not allow it to be measured (Derczynski, 2013). Thus, *after the rain fell* is not valid, unlike an expression such as *the day after the rain fell* which is centered around *day*, a measure of time. From this definition, we can establish a typology of temporal expressions. A temporal expression can most often be (see Derczynski et al., 2012 for a more thorough and complete typology):

**Absolute,** when a moment is totally explicit and unambiguous such as *Monday, October 6th, 2019*.

**Deictic**, when the moment of enunciation must be used to determine the moment to which the expression refers: *two weeks ago*. We can assume, for example, that the moment of enunciation is the moment when the text was written.

**Anaphoric,** when the moment of enunciation is distinct from the moment when the reference is made, when a person is telling a story in the past tense for example (*that evening*). The moment of enunciation (the moment when she tells the story) is not enough, it is necessary to determine the moment of reference within her story.

Given this typology, we can more easily identify in a text what we will call temporal expressions and what our questions will focus on. Here is an overview of what our system will have to process (text extracted from the WikiWars corpus (Mazur, Pawel, and Robert Dale (2010))):

*Royal flight to Varennes*

*(...)On the night of 20 June 1791 the royal family fled the Tuileries wearing the clothes of servants, while their servants dressed as nobles. However, the next day the King was recognised and arrested at Varennes (in the Meuse departement). He and his family were paraded back to Paris under guard, still dressed as servants. From this time, Barnave became a counselor and supporter of the royal family.*

A temporal question answering system will have to be able to answer questions on the temporal expressions highlighted in yellow. We can distinguish different types of temporal questions.

*Questions which have a **literal** answer*. The answer is found literally in the text and the system will need to be able to select the appropriate passage, corresponding to the answer sought. Questions that would fit into this category would be: *When did the royal family flee from Paris? When was the king arrested? When did Barnave become counselor of the royal family?*

*Questions that require **inference***. The answer is not directly present in the text and the system will need to be able to identify the temporal information it will need before reaching a conclusion: *How long was the king away from Paris?* (He left on June 20 and was arrested the next day, so he was gone 2 days). *What was the date when the king was arrested?* (*the next day* corresponds in fact to June 21st since the previous day was the 20th)

In this paper, we limit ourselves to literal questions, but these examples already give us a glimpse of how complex temporal question answering can be. We will begin by presenting the methods generally used in question answering, explaining which method we preferred and why. We will then review state of the art corpora by presenting the SQuAD (Rajpurkar & al., 2016) and especially the WikiWars (Mazur & Dale, 2010) corpora, explaining how we combined WikiWars with the SQuAD approach to create our own temporal corpus. We will then detail our model and explain how the data is represented and which features were used, before finally presenting and discussing the results obtained.

## 2 State of the Art and Methods

Traditional information retrieval involves finding a short passage of text within a set of documents. A selection of relevant documents is first made, then these documents are subdivided into sections, paragraphs, or sentences. We focus only the second part of this task, which we adapt to the case of temporal question answering.

We decided to mainly use the **information extraction** method (vs the Knowledge-Base approach) which means using literal questions as we stated above, mainly because annotating the data is faster: when building the dataset, we can write the questions as they are without worrying about translating them into logical form. It is also possible to ask a third party to help build the dataset since all that is needed is to write a question and identify the answer within the text. These advantages make it possible to build a larger dataset more quickly. However, we are not opposed to the Knowledge-Base approach and we even think that combining the two approaches could be something to explore in the future.

| Answer type | Percentage | Example |
|---|---|---|
| Date | 8.9% | 19 October 1512 |
| Other Numeric | 10.9% | 12 |
| Person | 12.9% | Thomas Coke |
| Location | 4.4% | Germany |
| Other Entity | 15.3% | ABC Sports |
| Common Noun Phrase | 31.8% | property damage |
| Adjective Phrase | 3.9% | second-largest |
| Verb Phrase | 5.5% | returned to Earth |
| Clause | 3.7% | to avoid trivialization |
| Other | 2.7% | quietly |

Table 1: Classification of answer types for SQuAD

## 2.1 The SQuAD corpus

SQuAD (Rajpurkar & al, SQuAD: 100,000+ Questions for Machine Comprehension of Text, 2016) is certainly one of the most well-known corpora when it comes to question answering. It is a corpus developed for question answering by extraction (the answer is literally present in the text and must be extracted) and it is for this reason that we have chosen to analyze it more closely, and eventually to draw inspiration from the methodology used.

The corpus is composed of articles from the English Wikipedia divided into paragraphs. There are 536 articles, chosen among the 10,000 most popular articles. The popularity of an article was determined using Wikipedia's Internal PageRanks from Project Nayuki, a site that offers a variety of practical computer applications (https://www.nayuki.io/page/computing-wikipedias-internal-pageranks). The PageRank of a document is the probability that a visitor will arrive at that document after performing a uniform random web search (uniform random browsing). From this selection, individual paragraphs are then extracted from each article, with those under 500 characters being eliminated.

As stated above, a response is equivalent to a passage extracted from a paragraph, which greatly simplifies the annotation of the data, and explains how the corpus can be so large (23,215 paragraphs in all). Indeed, the questions were produced through intensive crowdsourcing. It is important to note that any type of question is valid, as long as a passage of text can be selected to answer it. SQuAD is therefore not a corpus that is particularly adapted to questions on temporal expressions, and this is one of the limitations of this corpus, as far as we are concerned.

Indeed, when looking at the types of responses contained in the corpus in Table 1 and the percentages they represent, few dates (proportionally) are highlighted as responses. Only 9% of the answers are dates, which shows that they are not the primary concern of the corpus. Moreover, these statistics, given by the authors, make it difficult to determine to what extent other types of temporal expressions (defined in the introduction) are present (durations, deictic expressions, anaphors, etc.)

## 2.2 The WikiWars corpus

On the other hand, WikiWars: A New Corpus for Research on Temporal Expressions, (2010) is better suited to our task in terms of content. The corpus was developed from 22 English Wikipedia documents that describe the historical courses of wars. The authors of the corpus searched Google for these two phrases: "most famous wars in history" and "biggest wars". They found a page describing the 10 most famous wars in history and a page describing the 20 most important wars of the 20th century. They then combined these two lists, eliminated duplicates and searched Wikipedia for articles about these wars. Here is an example of a paragraph from the WikiWars corpus:

> *On <TIMEX2 val="1791-06-20TNI">the night of 20 June 1791</TIMEX2> the royal family fled the Tuileries wearing the clothes of servants, while their servants dressed as nobles. However, <TIMEX2 val="1791-06-21">the next day</TIMEX2> the King was recognised and arrested at Varennes (in the Meuse departement)*

We have highlighted the temporal expressions as well as the TIMEX2 tags that surround them. The TIMEX2 annotation scheme (Ferro et al., 2005) allows us to associate a temporal value with the expression in question, which could be leveraged in further work involving inference questions (expressions such as "the next day" have dates associated with them (1791-06-21), which would allow for questions such as "What day was it when the King was arrested ?" to have a more precise answer than simply "the next day"). However, given our focus on purely extracting responses from the text, the *val* attribute was not used.

WikiWars holds a greater number of references to the distant past and the temporal structures of the texts are more elaborate than those found in SQuAD. Furthermore, the number of temporal expressions per document is higher than other popular temporal corpora (121.41 timex/doc vs. 7.73 for the ACE corpus (Doddington, 2005)). This therefore makes it a better fit for our task. However, unlike SQuAD, it is not annotated for question answering. We believe that the combination of these two types of corpora has not yet been sufficiently explored and we therefore created a dataset that addresses this shortcoming.

## 3 Data and Model

### 3.1 Using the SQuAD approach on the WikiWars corpus

We combined the SQuAD approach with the WikiWars corpus, in order to test the extraction method on a corpus suitable for the study of temporal expressions. WikiWars is not annotated with question-answer pairs, so we augmented this dataset to suit our task, by breaking the text into paragraphs, like the SQuAD documents, and adding a list of questions and answers under each of them, using XML tags.

In order to enrich our dataset considerably, we decided that three questions would be associated with each temporal expression. As well as providing a larger training set, this meant that copying elements from the text to formulate the questions (and therefore simplifying the task of finding and extracting the answer for the model) was necessarily limited since the questions could not resemble each other exactly, as illustrated in the following example:

*On September 1, 1939 Germany and Slovakia (...) attacked Poland and World War II broke out.*

This extract could have as associated questions: *When did World War II break out? What day was it when WWII started? When was Poland attacked?*

Rephrasing makes it more difficult for the model to determine which part of the paragraph the question is about. In the example above, the first question uses information at the end of the sentence while the answer is at the beginning; the second question uses *started* instead of *broke out* and synthesizes *World War II* into *WWII*; the last question is in the passive voice, thus reversing the order of the words found in the text. The efficiency of the model is therefore tested by using such examples, especially since some paragraphs can be quite long (the longest ones contain approximately 300 words).

Given the amount of question-answer pairs to annotate (approximately 6000) and the straightforwardness of the annotation task, the annotations were performed manually by 2 bilingual annotators (French and English). In the annotation protocol, the annotators were provided with a presentation of the WikiWars corpus and with an explanation of our aim in creating this corpus. They were also provided with several examples of annotated paragraphs and guidelines which insisted on reformulating the text when writing the questions and on finding various formulations.

Not all temporal expressions were taken into account. Indeed, it was sometimes difficult to ask coherent questions which took these expressions as answers. Given that the priority of our task was to have logical and coherent questions that a user could ask, we felt that if questions became too artificial (to accommodate a particular temporal expression as an answer), then they should not appear in our dataset.

For example, the adjective *former* sometimes caused problems. Although we can see how this adjective can provide useful temporal information, formulating a question which has this specific word as an answer does not sound natural, as can be seen in the following example:

*(...)Republican former vice president Richard Nixon.*

*What vice president was Nixon? => (?)Former*

We therefore asked our annotators to leave the field blank if they felt that a question might be difficult to phrase and proof-read their annotations.

In total, our corpus contains 702 paragraphs (paragraphs without temporal expressions were not counted) and 6120 question-answer pairs, which were annotated in approximately a month and a half. By comparison, SQuAD has around 23,000 paragraphs and 107,000 question-answer pairs. Although the amount of data is not as large, it is much more specific and only targets temporal information.

The dataset can be acquired and used for other experiments by contacting armand.stricker@universite-paris-saclay.fr or benoit.crabbe@linguist.univ-paris-diderot.fr.

### 3.2 Model

Neural networks are particularly well suited for extracting answers from a text and it is this approach that we have chosen. Indeed, to try to answer the question, the model will try to find similarities between the words in the question and the words of the paragraph by comparing their respective distributional representations. We chose to use recurrent neural networks, since they are ideal to encode the information contained in a sequence.

We implemented a model inspired by the Document Reader component of the DrQA system designed by Chen & al. (2017), a system that allows a user to search for a document and then select a passage within it. Thus, a question is composed of $l$ tokens :

$$Q = \{q_1, q_2, \ldots, q_l\} \quad (1)$$

and a paragraph is composed of $m$ tokens :

$$P = \{p_1, p_2, \ldots, p_m\} \quad (2)$$

**Paragraph encoding** For each word in the paragraph, we first create a vector representation which is the concatenation of 4 components, all of which are intended to try to draw the model's attention to certain words in the paragraph, rather than others. Here are the functions that translate these different features:

*Word embeddings* - We first use 300-dimensional GloVE pre-trained embeddings (Pennington & al., 2014) to obtain the embedding of a word $p_i$:

$$f_{embedding}(p_i) = \boldsymbol{E}(p_i) \quad (3)$$

*Exact match* - This function creates two features: the fact that a word $p_i$ is identical in the question and in the paragraph, and the fact that the lemmatized forms of the token are also identical:

$$f_{exact\_match}(p_i) = \mathbb{I}(p_i \in Q) \quad (4)$$

*Token Features* - We encoded the various characteristics of a token $p_i$, namely its grammatical category (POS, part of speech), whether it is part of a named entity (NER, named entity recognition), and the TF-IDF (term frequency - inverse document frequency):

$$f_{token}(p_i) = concat(POS(p_i), \ldots) \quad (5)$$

To obtain the POS of a word, we used the automatic nltk POS-tagger (https://www.nltk.org/book/ch05.html). To obtain named entities, we used spaCy (https://spacy.io/usage/linguistic-features#named-entities), who trained its algorithm on OntoNotes (Weischedel & al., 2011). The algorithm is capable of identifying a range of entities, and most importantly dates. As for the TF-IDF, this measure allows us to weigh the frequency of the token $p_i$ by seeing if it is present in other examples. The more the token is present in the corpus, the lower its weighted frequency will be.

*Aligned embedding of the question (attention mechanism)* - Finally we added an attention vector: often, in addition to encoding the exact match, question answering systems use an attention mechanism to represent in a more sophisticated way the similarity between a passage and a question, for similar but non-identical words like *flight* and *plane* for example. The vector is supposed to reflect the proximity between the token and the words in the question. We use a weighted similarity function where $p_i$ represents the queries and $q_j$ the keys:

$$f_{aligned}(p_i) = \sum_j a_{i,j} \boldsymbol{E}(q_j) \quad (6)$$

The attention weight $a_{i,j}$ encodes the similarity between the token $p_i$ and each word $q_j$ in the question. This attention weight can be calculated as the dot product between the functions $\alpha$ of the question words' embeddings and the paragraph's, where $\alpha$ can be a simple feed forward neural network:

$$a_{i,j} = \frac{\exp\left(\alpha(\boldsymbol{E}(p_i))^T . \alpha(\boldsymbol{E}(q_j))\right)}{\sum_{j'} \exp\left(\alpha(\boldsymbol{E}(p_i))^T . \alpha(\boldsymbol{E}(q_{j'}))\right)} \quad (7)$$

We concatenate all these feature vectors to obtain a vector representation for each token in the paragraph:

$$\widetilde{p}_i = concat(f_{embedding}(p_i), ...) \quad (8)$$

Finally, each $\widetilde{p}_i$ is passed through an RNN so as to obtain a final $p'_i$ for each token:

$$\{p'_1, p'_2, ..., p'_m\} = RNN(\{\widetilde{p}_1, \widetilde{p}_2, ..., \widetilde{p}_m\}) \quad (9)$$

**Question encoding** The question encoding is similar to the paragraph encoding but is simpler because not as many features are used to represent each token in the question. Pre-trained embeddings such as GloVE (Pennington & al., 2014) are used to obtain the vector representation $\widetilde{q}_i$ which will be transmitted to the RNN (an LSTM (Hochreiter & al., 1997) in our case). We do not create any other features for the tokens in the question:

$$\widetilde{q}_i = f_{embedding}(q_i) \quad (10)$$

The sequence is encoded and we output the hidden representations of the network:

$$\{q'_1, q'_2, ..., q'_l\} = RNN(\{\widetilde{q}_1, \widetilde{q}_2, ..., \widetilde{q}_l\}) \quad (11)$$

These vector representations are then combined through a weighted sum to produce a single vector **q** which represents the question:

$$q = \sum_j b_j q'_j \quad (12)$$

The weight $b_j$ is a measure that reflects the relevance of each word in the question and can be learned from a weight vector **w**:

$$b_j = \frac{\exp(w.q'_j)}{\sum_{j'} \exp(w.q'_{j'})} \quad (13)$$

**Prediction of a span -** Once the two previous steps are completed, we obtain an embedding of the question **q** and a vector representation for each token of the paragraph $\{p'_1, p'_2, ..., p'_m\}$. We then train two different classifiers: one to calculate the probability $P_{start}(i)$ that a word $p_i$ marks the beginning of the answer and one to calculate $P_{end}(i)$. We use a bilinear attention layer as a similarity function, where $W_{start}$ and $W_{end}$ are matrices of learned weights:

$$P_{start} \exp(p'_i W_{start} q) \quad (14)$$

$$P_{end} \propto \exp(p'_i W_{end} q) \quad (15)$$

One way to determine which passage is the answer is to take the word with the highest start probability and the one with the highest end probability, in order to deduce that the words in between are part of the answer. This is obviously not the only way to proceed, and it would also have been possible to build a model that goes through the text predicting whether or not a word is part of the answer.

We use cross-entropy loss as our loss function, where the argmax position of the vectors containing the start and end probabilities is compared to the indices of the gold labels in the example in order to update the model's parameters.

### 3.3 Training

We split our documents into three csv files: train, dev and test. The paragraphs of each document were split into three temporary datasets with the following proportions: 80%, 10%, 10%. Since each paragraph contains several questions, distributing the data in this way and not per question, avoids that a paragraph be present in both the dev and the train dataset, which would distort the results.

## 4 Results and discussion

### 4.1 Metrics

We used several metrics to evaluate the performance of our model. The model systematically predicts a start and end token, so we evaluated: *the percentage* of tokens at the beginning of an answer correctly predicted, *the percentage* of correctly predicted tokens at the end of a response, *the mean* of the two previous measures, *the percentage* of whole passages correctly predicted (the start token and the end token are correct, it is an exact match).

The table below shows our results, on the development and test sets:

| Our Model | Dev | | Test | |
|---|---|---|---|---|
| | Mean | Exact match | Mean | Exact Match |
| | 61.5 | 52.6 | 54.7 | 41.4 |

Table 2: Results on dev and test

We see that our model manages to predict correctly about 50% of the passages on the dev set. We also notice that the performances are lower on the test corpus. It is difficult to evaluate why, except to consider the fact that the corpus may present examples that are too different from those encountered during training. We will take a closer look at the errors made by our model in the next section.

### 4.2 Ablation Analysis

We also conducted an analysis of the features used to encode the paragraphs on the development set by ablation, as shown in the table below:

| Features | F-measure |
|---|---|
| Full | 52.6 |
| | |
| No NER | 48.3 |
| No POS | 50.1 |
| No TF-IDF | 51.5 |
| No $f_{token}$ | 43.5 |
| No $f_{aligned}$ | 46.3 |
| No $f_{exact\_match}$ | 49.4 |
| | |
| No $f_{aligned}$ et $f_{exact\_match}$ | 35.11 |
| No $f_{aligned}$ et NER | 46.2 |
| No $f_{exact\_match}$ et NER | 43.3 |

Table 3: Ablation analysis for the features used

We notice that the f-measure (which boils down to the percentage of exact matches in this case) does not drop much (3%) when we remove the exact match, traditionally a very important feature for general question answering. We also notice that the removal of the NER (named entity recognition) does not lead to a drastic drop either (4.3%) even if the algorithm used (https://spacy.io/usage/linguistic-features#named-entities) allows us to identify temporal expressions quite reliably. This means that even when the possible answers are indicated in the paragraph, our model does not rely only on this feature to find the answer, and that it is not enough to simply extract the temporal expressions of a paragraph to find the right answer. The features that account for the link between the question and the text are therefore of great importance.

What is interesting to note is the interaction of the different features with each other. Indeed, the individual ablations of attention ($f_{aligned}$) and of $f_{exact\_match}$ generate relatively small decreases (6.3% and 3.2%). But when we remove both features simultaneously, the performance of the model drops drastically: by 17.4%. We can conclude that these two features play a similar but complementary role and that they are quite essential in the search for the passage in the paragraph, allowing to identify the context within which to look for the answer.

Nevertheless, not all features interact with each other. In another case, the decrease is additive: the simultaneous removal of the information on the grammatical category (POS), on the named entities (NER) and of the TF-IDF generates a score of 43.5% ("No $f_{token}$" in table 3, where $f_{token}$ is a concatenation of the features mentioned). This score corresponds approximately to the sum of the individual losses caused by each feature (respectively - 2.5%, - 4.3% and - 1.1%, which would result in a score of 44.7%). However, 43.5% is indeed slightly lower than 44.7%, so there must be some interaction. We also tested the interaction between named entity recognition and contextualization features $f_{exact\_match}$ and $f_{aligned}$, but the drop in performance was not significant.

### 4.3 Qualitative Analysis of Model Inference

In the following examples, "(START)" and "(END)" indicate the boundaries of the expected response, while "****_" and "_****" indicate the start and end tokens predicted by our model. They appear only when the prediction is wrong. The examples were taken directly from the output of the model.

Overall, our model predicts almost only temporal expressions. When there is an error, the answer is often not the temporal expression expected or turns out to be incomplete, as we will see through the following series of examples.

**Correctly handled cases -** The most favorable case for predicting a correct answer is when the paragraph is relatively short and the question is formulated in such a way that the context of the answer is easy to identify and relatively close to the answer. The model also performs better when the choice of temporal expressions within the paragraph is limited, as in the following example:

> when did emperor charles i attempt secret negotiations with <unk> ?
>
> in (START)1917(END) , emperor charles i of austria secretly attempted separate peace negotiations with <unk> , with his <unk> ' s brother <unk> in belgium as an intermediary , without the knowledge of germany . (...)

This example is quite short and *1917* is the only temporal expression. We have highlighted the shared passages in the question and the paragraph. The words are exact matches here, except for "attempt" and "attempted" although they still have the same lemma.

**Partially correct answers** - The fact that our model is trying to predict a start and end token means that several temporal expressions are sometimes combined by our model, as in the example below:

> when was south africa invaded by german troops ?
> some of the first clashes of the war involved british , french and german colonial forces in africa . on ****_7 august , french and british troops invaded the german <unk> of <unk> . on (START)10 august(END) german forces in south - west africa attacked south Africa; sporadic and fierce fighting continued for the remainder of the war .

Our model predicted the wrong start token but the correct end token. The predicted start token is indeed a date but is not part of the correct temporal expression. It is as if the model had tried to combine *7* and *august*, surely because the context around these two terms fits the question well (we have highlighted the exact matches around the two temporal expressions). Modeling the proximity between temporal expressions within a paragraph could therefore be considered during future experiments. Nevertheless, we see the importance of the model's ability to contextualize correctly, and this type of error can even lead to completely inconsistent answers, where the end token is predicted to be before the start token.

**Embeddings** - We also observed a lack of expressiveness in the embeddings used. When a token in the question had a synonym in the paragraph, this did not always help guide the model. Indeed, the model seemed sometimes disoriented when a synonym for a word near the answer was used, resulting in an incorrect prediction.

**Year, day, and season -** Our model seemed to be able to recognize these key words in the questions and understood what format the answer was expected to have. This was especially apparent when we compared the answers for "when" questions (which were note as precise) with "what year", "what day", "which season" questions for the same paragraph.

## 5 Conclusion

In this work, we chose to simplify the temporal question answering problem and limit our work to literal questions. We were thus able to apply an extractive approach with some success.

Indeed, this method allowed us to create a rather large dataset, annotated by hand, over a rather short period of time (only about a month and a half). Thanks to this, we were also able to devote time to the state-of-the-art implementation of a deep machine learning model, the results of which demonstrated some of this model's strengths.

Nevertheless, it is not clear that the model is capable of understanding the underlying structure of the text it is dealing with. When the model has to choose between several temporal expressions, searching for the tokens closest to the answer does not always lead to the right prediction.

Moreover, our model is certainly not capable of making inferences and determining, for example, the date to which a deictic temporal expression, such as "this day", refers. In future work, we propose to broaden the definition of a temporal question in order to be able to deal with a larger variety of questions, especially inferential questions.